\documentclass[conference]{IEEEtran}
\IEEEoverridecommandlockouts

\usepackage[utf8]{inputenc}  
\usepackage{cite}
\usepackage{amsmath,amssymb,amsfonts}
\usepackage{algorithmic}
\usepackage{graphicx}
\usepackage{textcomp}
\usepackage{xcolor}
\usepackage{url}
\usepackage{booktabs}
\usepackage{comment}
\usepackage{caption}  
\usepackage{subcaption}  
\usepackage{soul}
\usepackage{multirow}
\newtheorem{definition}{Definition}

\def\BibTeX{{\rm B\kern-.05em{\sc i\kern-.025em b}\kern-.08em
    T\kern-.1667em\lower.7ex\hbox{E}\kern-.125emX}}
\newcommand{\chiaraR}[1]{{\color{magenta}{#1}}}

\makeatletter
\newcommand{\linebreakand}{%
  \end{@IEEEauthorhalign}
  \hfill\mbox{}\par
  \mbox{}\hfill\begin{@IEEEauthorhalign}
}
\makeatother

\begin{document}

\title{Urban Region Embeddings from\\Service-Specific Mobile Traffic Data\\
}

\author{\IEEEauthorblockN{Giulio Loddi}
\IEEEauthorblockA{
\textit{IMT Lucca}\\
Lucca, Italy \\
giulio.loddi@imtlucca.it}
\and
\IEEEauthorblockN{Chiara Pugliese}
\IEEEauthorblockA{\textit{ISTI-CNR, University of Pisa} \\
Pisa, Italy \\
chiara.pugliese@isti.cnr.it}
\linebreakand
\IEEEauthorblockN{Francesco Lettich}
\IEEEauthorblockA{\textit{ISTI-CNR} \\
Pisa, Italy \\
francesco.lettich@isti.cnr.it}
\and 
\IEEEauthorblockN{Fabio Pinelli}
\IEEEauthorblockA{\textit{IMT Lucca} \\
Lucca, Italy \\
fabio.pinelli@imtlucca.it}
\and
\IEEEauthorblockN{Chiara Renso}
\IEEEauthorblockA{\textit{ISTI-CNR} \\
Pisa, Italy \\
chiara.renso@isti.cnr.it}
}

\maketitle

\begin{abstract}
With the advent of advanced 4G/5G mobile networks, mobile phone data collected by operators now includes detailed, service-specific traffic information with high spatio-temporal resolution. In this paper, we leverage this type of data to explore its potential for generating high-quality representations of urban regions. To achieve this, we present a methodology for creating urban region embeddings from service-specific mobile traffic data, employing a temporal convolutional network-based autoencoder, transformers, and learnable weighted sum models to capture key urban features.
In the extensive experimental evaluation conducted using a real-world dataset, we demonstrate that the embeddings generated by our methodology effectively capture urban characteristics. Specifically, our embeddings are compared against those of a state-of-the-art competitor across two downstream tasks. Additionally, through clustering techniques, we investigate how well the embeddings produced by our methodology capture the temporal dynamics and characteristics of the underlying urban regions.
Overall, this work highlights the potential of service-specific mobile traffic data for urban research and emphasizes the importance of making such data accessible to support public innovation.

\end{abstract}


\section{Introduction}
\label{sec: intro}

Mobile phone activity data is a well-established and widely explored type of mobility data used in various applications, including mobility, health, socio-economic, and demographic studies. 
In the past years, mobile phone data was typically studied in the form of Call Detail Records (CDRs), which capture users' connections to cell towers during calls or messaging activities. However, this type of data is often sparse and irregular, limiting its potential for broader and more scalable applications.

With the rise of 4G/5G cellular networks, mobile phone usage has shifted towards extensive use of data services, such as \textit{mobile applications}, which generate massive volumes of data traffic. 
The information related to the data traffic volume generated by these services can offer rich spatio-temporal details and insights into the characteristics of the underlying urban regions. 
To this end, in this work, we consider the NetMob 2023 dataset \cite{martínezdurive2023netmob23}, 
which provides detailed data on mobile traffic volume across multiple data services. Orange, the mobile operator providing the dataset, recorded upload and download traffic for 68 different mobile applications across 20 major French cities.
More specifically, we focus on studying this type of service-specific mobile traffic data in the context of \textit{urban region representation}. 

Urban region representation is a broad research field encompassing geography, urban planning, architecture, social sciences, and cultural studies. It plays a key role in smart city development and aims to represent the mobility and dynamics of populations. With the increasing availability of different types of data collection devices monitoring urban spaces, there is growing interest in using new and dynamic data sources to improve 
urban region representation.
%
%
Urban region representation is generally studied within the broader field of urban computing, a discipline that integrates computing, data analysis, and urban planning to address complex problems in urban environments \cite{zheng2014urban}.
%
Different task-specific methods have been developed in the urban representation field,
such as for predicting economic growth \cite{hui2020predicting} or forecasting air quality \cite{zheng2013u}.
 
Urban environments are inherently complex, characterized by multiple, interacting dynamics. To address this complexity, specially tailored methods and datasets are often required. Mobile phone data offers extensive and uniform coverage across urban areas, thus potentially addressing some challenges associated with multi-modal data.

Going beyond such ad-hoc methods for specific application tasks, several recent research efforts have focused on \textit{urban region representation learning}.  This approach involves using deep learning techniques to create task-independent embeddings of urban regions 
\cite{li2023urban,wang2020urban2vec}. 
Embeddings are low-dimensional vectors that capture the salient features of urban regions. One advantage of embeddings is their versatility: they can be used as input for models employed in downstream tasks to predict some characteristics of urban regions, e.g., population density, land use, and demographic indicators.
Traditionally, studies in this domain have generated urban region embeddings using data sources commonly considered in the literature, such as Points of Interest (POIs), land use, and satellite imagery.

Different from the existing literature, in this work, we aim to exploit service-specific mobile traffic volume data to capture and study the characteristics of urban regions.
%
%
Furthermore, we intend to take advantage of the characteristics of the NetMob 2023 dataset to conduct temporal analyses that are not possible with static datasets based on, e.g., POIs or land use data. Indeed, we argue that service-specific traffic volume data is very valuable for generating urban areas representations because it reflects the widespread use of mobile devices, generating continuous, fine-grained spatio-temporal data. Different services represent various human activities, e.g., commuting, socializing, and entertainment, allowing for dynamic insights into urban behavior. Such data can also reveal socio-economic patterns, as usage intensity and service types vary across different regions. Moreover, unlike static datasets, mobile traffic volume data continuously changes, capturing the evolving dynamics of urban regions. 

We pause one moment to make an important observation: despite the availability of aggregated service traffic data, as well as the spatio-temporal capillarity with which it is gathered across urban regions, such data is often held by private companies, hence limiting its availability for public research and innovation (such as in urban planning). Demonstrating the effectiveness of this data in producing high-quality urban representations could encourage these companies to share their data with researchers for the development of novel methods and with stakeholders for business or public benefit.

\vspace{0.3em}
In summary, the fundamental \textbf{research question} we aim to answer is: ``\textit{can service-specific mobile traffic volume data be used to extract representations of urban region that are effective in capturing the regions' salient features?}"

\vspace{0.3em}
To answer this question, we first formally introduce some fundamental notions 
and then the problem we want to address (Section \ref{sec: preliminaries}). Subsequently, we introduce the methodology we propose to tackle the problem, i.e., generate embeddings of urban regions from service-specific mobile traffic volume data (Section \ref{sec:methodology}). Finally, in the experimental evaluation (Section \ref{sec:experiments}), we first apply our methodology to the NetMob 2023 dataset, thus generating urban region embeddings for selected cities, and then analyze these embeddings from multiple points of view.

Specifically, we first show that service-specific mobile traffic data can be effectively used to generate urban region embeddings, outperforming those derived from more traditional data sources. We evaluate our findings through two supervised tasks and one unsupervised task, comparing them with RegionDCL \cite{li2023urban}, a state-of-the-art method. Notably, the evaluation of the unsupervised task underscores the ability of our methodology to generate embeddings that capture rich urban information.
Finally, the qualitative analysis shows that the embeddings generated by our methodology capture the temporal dynamics of urban regions, thus providing a basis for analyzing changes over time -- for instance, differences in transportation needs across various intervals, or insights into traffic patterns during weekdays, weekends, and holidays.



\section{Related work}
\label{sec:related}

The use of mobile phone data for urban analysis has been extensively studied over the past two decades \cite{ratti, Mobilesmartcities15, surveymobilephone23}. However, the use of service-specific mobile traffic data is relatively new, primarily due to recent advancements in 4G/5G technology and the limited availability of open datasets. The release of the NetMob 2023 dataset, though restricted to challenge participants \cite{martínezdurive2023netmob23}, has encouraged researchers to explore the analysis of such data and its potential applications \cite{Netmob23abstract}.
To the best of our knowledge, neither the NetMob 2023 dataset nor similar datasets on service-specific mobile traffic have been applied to urban region representation or related downstream tasks.  

From the point of view of urban representation, several studies have explored the generation of unsupervised representations of urban regions using contrastive learning-based approaches. For instance, \cite{wang2020urban2vec} uses street view images and POI data to learn neighborhood embeddings. In \cite{li2023urban}, features extracted from building footprints are combined with POI data, producing representations at both the building group-level and coarser region-level aggregations to account for different urban partitions. Satellite imagery has also been used to extract urban features, either in combination with POI data  \cite{xi2022beyond} or with LLM-generated textual descriptions  \cite{yan2024urbanclip}. Similar approaches employ mobility flows for urban region representation \cite{wang2017region, jenkins2019unsupervised}. Other recent methods use multi-view contrastive prediction to compute region embeddings \cite{Regionembeddingmultiview24}.



Similarly to our work, these approaches employ contrastive learning for region representation. 
However, to the best of our knowledge, no approach in the literature considers service-specific mobile traffic data for urban region embeddings and analysis. In this regard, the methodology we propose in this work provides a complementary perspective to existing state-of-the-art methods.

%
Finally, a common theme in the literature is the integration and use of multi-modal data to produce high-quality urban representations. We believe that exploiting service-specific mobile traffic 
data offers a unique way to capture the variety of people behaviors and activities within each urban region.



\section{Preliminaries and problem definition}
\label{sec: preliminaries}

In this work, we consider the \textit{service-specific mobile traffic data} measured by a mobile operator across an urban area of interest at some spatio-temporal granularity. From this, our goal is to extract, discover, and analyze patterns to gain insights into the spatio-temporal dynamics and characteristics of \textit{urban regions}.
We begin tackling the problem by introducing the geographical unit from which the mobile operator measures the traffic volume data. This effectively materializes a tessellation over the urban area and can be seen as the source of information we employ to study it. 
We refer to this tessellation as the \textit{mobile traffic cell (MTC)} tessellation.

\vspace{0.3em}
\begin{definition}[\textsc{Mobile Traffic Cell tessellation}]
\label{def: mtc}
Consider a mobile operator that collects mobile traffic volume data for an urban area of interest $A$ at a certain spatio-temporal granularity.
To achieve this, the operator tessellates $A$ into a set of cells $\mathcal{C}$ whose shape and size are determined by the operator.
Within each cell, the operator tracks the mobile traffic volume generated by a set of services $\mathcal{S}$ over a specified time period. 
We name the cells in $\mathcal{C}$ as \textit{mobile traffic cells} (MTCs).
\end{definition}

\vspace{0.2em}
As a concrete example, the NetMob 2023 dataset uses a grid tessellation of $(100m)^2$ square cells which corresponds to the MTC tessellation, with data collected at this specific spatial resolution. Therefore, the tessellation used in the dataset corresponds to the MTC tessellation defined in Definition \ref{def: mtc}.

Next, we define how the mobile traffic volume for a specific service $s \in \mathcal{S}$ in a particular MTC $c \in \mathcal{C}$ is represented as a time series. 

\vspace{0.3em}
\begin{definition}[\textsc{Cellular Time Series (CTS)}]
\label{def: stvts}
Let $T_{c,s}\in \mathbb{R}^k$ be the time series representing the mobile traffic volume recorded in the MTC $c \in \mathcal{C}$ generated by a service $s \in \mathcal{S}$ over $k$ timesteps. We name such time series as Cellular Time Series (CTS).
\end{definition}

\vspace{0.2em}
Going back to the previous example, in the NetMob 2023 dataset, for each MTC cell mobile traffic volume is measured for various services, including streaming services like YouTube, gaming services like Fortnite, and email services like Gmail. Accordingly, for each service $s \in S$ and MTC $c \in \mathcal{C}$ pair, the dataset provides a CTS spanning 2.5 months, from March 16th, 2019 to May 31st, 2019, with a sampling frequency of 15 minutes, resulting in $k=7392$ samples.

\vspace{0.3em}
In general, MTC tessellations used by mobile phone operators differ from urban area subdivisions used by most common stakeholders, e.g., urban planners, local and public transport authorities, etc.
For example, an urban area can be divided into functional regions (e.g., residential, commercial, industrial), regions of historical significance (e.g., neighborhoods), or regions defined by statistical properties (e.g., IRIS cells in France, Census Block Groups in the US). 
Therefore, to provide meaningful information to final users from data provided by mobile phone operators, we need to introduce the notion of \textit{target tessellation}.


\vspace{0.3em}
\begin{definition}[\textsc{Target tessellation}]
\label{def: target partitioning}
Given an urban area of interest $A$, a target tessellation divides $A$ into a set of non-overlapping regions $\mathcal{R} = \{R_1,R_2,\ldots, R_N\}$. 
\end{definition}

\vspace{0.3em}
At this point, we are ready to state the problem definition. 

%
%
\vspace{0.3em}
\begin{definition}[\textsc{\textbf{Problem Definition}}]
\label{def: prob def}
Given an urban area of interest $A$, discretized with a MTC tessellation $\mathcal{C}$, and a set of  cellular time series $\mathcal{T}$, relative to the usage of a set of services $\mathcal{S}$ over $\mathcal{C}$, and a target tessellation $\mathcal{R}$, we aim to 
generate from $\mathcal{T}$ high-quality embeddings for the regions in $\mathcal{R}$.
\end{definition}

\vspace{0.3em}
In the next section, we propose a 
methodology 
to tackle the problem of exploiting service-specific mobile traffic data to compute region embeddings and analyze the region characteristics and dynamics.

\section{Methodology}
\label{sec:methodology}

\begin{figure*}[ht]
  \centering
  \includegraphics[width=0.9\textwidth]{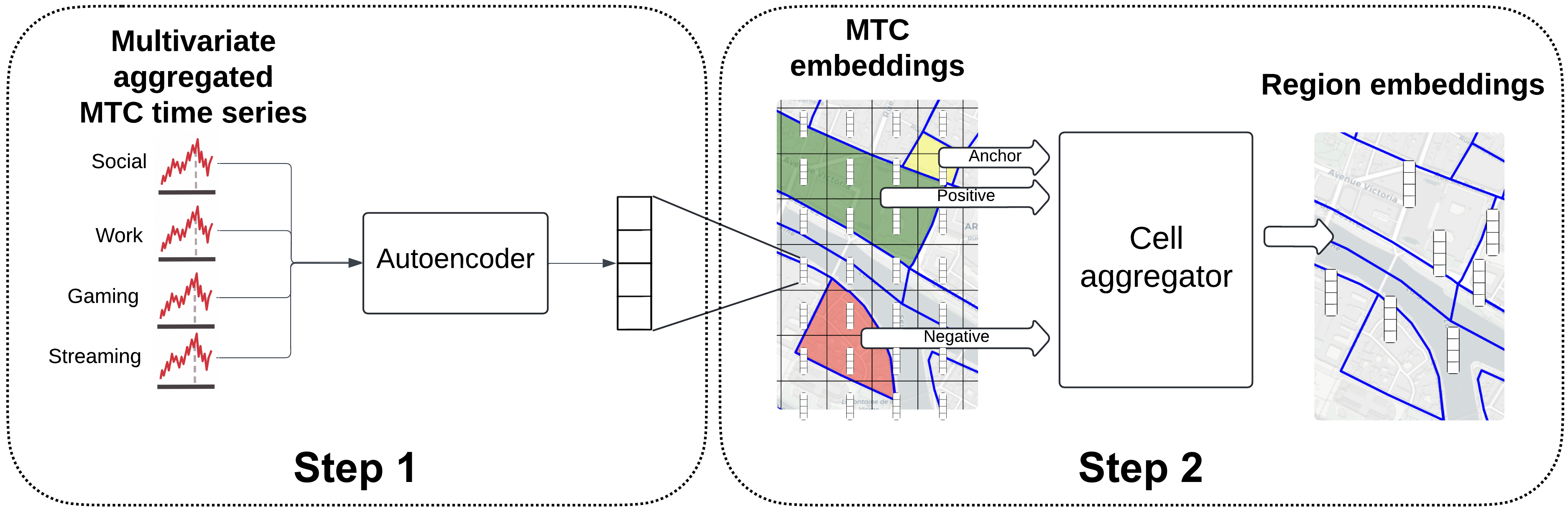} 
  \caption{Summary diagram of the methodology: multivariate aggregated MTC time series are inputted to an autoencoder (\textbf{Step 1}), which outputs low-dimensional MTC embeddings. These embeddings are then given as input to a Cell Aggregator (\textbf{Step 2}), which acts as an MTC aggregator to generate the final urban region embeddings. The aggregation is performed under a contrastive learning objective, which captures spatial dependencies between regions.
  }
  
  \label{fig:architecture}
\end{figure*}

In this section, we present the methodology designed to address the problem introduced in Definition \ref{def: prob def}, Section \ref{sec: preliminaries}. Broadly speaking, the approach is structured into two main steps. 
In the first step, we use the multivariate aggregated MTC time series to train an autoencoder, which generates the MTC embeddings. In the second step, these MTC embeddings are used to generate urban region embeddings corresponding to the target tessellation through a cell aggregator. This is done in a fully unsupervised setting using a contrastive learning task.

%

As mentioned earlier, mobile operators typically measure the traffic volume of many services, resulting in multiple CTSs for each MTC. However, some services might not be relevant, while others can be grouped into broader categories that serve as proxies for capturing urban region features. Since our focus is on understanding urban activities rather than individual services, we apply some \textit{preprocessing} operations. Specifically, we aggregate CTSs related to similar services to form multivariate time series for each MTC. For instance, for each MTC, we combine the traffic volume of YouTube, Netflix, and Prime Video into a single multivariate time series representing the ``streaming" category. Additionally, we perform temporal downsampling of the CTSs to match a desired sampling interval. If the data is already at the required interval, this step can be skipped. Downsampling is achieved by summing the values within each new time bin.

Each step of the methodology is detailed below.

\vspace{0.3em}
\noindent \textbf{Step 1 -- 
Generating MTC embeddings from multivariate aggregated time series.}
The first step involves generating \textit{MTC embeddings} from the multivariate aggregated time series computed in the first step. These embeddings are low-dimensional vector representations of the MTCs, designed to capture their most essential features from the time series data.
To generate the embeddings, we first train an autoencoder, a specific type of deep learning model, on the multivariate aggregated time series. Once trained, the autoencoder is used to generate the embeddings.

An autoencoder is a neural network designed to learn efficient data representations by progressively compressing the input into a lower-dimensional space (also known as the bottleneck layer) and then reconstructing the original input from this compressed version. The bottleneck layer represents the embedding space, where the model attempts to capture the most important features of the data. 
Training an autoencoder is a form of unsupervised learning, typically framed as a reconstruction task in which the model aims to minimize the error between the original data and its reconstructed version generated from the embeddings.

There are various types of autoencoders for time series available in the literature. In our work, we employ the one based on temporal convolutional networks \cite{thill2020time} (TCN-Autoencoder in Figure \ref{fig:architecture}). Although we also evaluated LSTM-based and GRU-based autoencoders, the TCN-Autoencoder achieved the lowest reconstruction loss in preliminary experiments, hence making it our preferred choice. Finally, we opted for a TCN-based autoencoder over more recent state-of-the-art ones because it presents a reasonable balance between performance, utility, and complexity, in the context of the problem we are addressing.

\vspace{0.3em}
\noindent \textbf{Step 2 -- 
Generating region embeddings from MTC embeddings.} 
The second step involves generating embeddings for urban regions, defined by the target tessellation $\mathcal{R}$ (see also Definition \ref{def: target partitioning} in Section \ref{sec: preliminaries}), from the MTC embeddings. This is achieved by using some aggregator $g$.
The aggregator is trained on a triplet loss function. In order to compute the loss, we first need to define the notions of anchor region, positive region, and negative region. An anchor region $RA \in \mathcal{R}$ is the region for which we want to compute an embedding. A positive region $RP \in \mathcal{R}$ is sampled from the neighboring regions of $RA$, while a negative region $RN \in \mathcal{R}$ is sampled from non-neighboring regions. The idea is that the embedding of $RA$ should be more similar to the embeddings of its neighboring regions (positive regions) than to those of more distant regions (negative regions). This strategy is in line with recent literature \cite{li2023urban,wang2020urban2vec} that follows Tobler’s first law of geography \cite{tobler1970computer}, which states that ``\textit{everything is related to everything else, but near things are more related than distant things}".

Accordingly, for each region $R$, we define $w_R$ as the set of MTC embeddings associated with the MTCs intersecting that region. 
Thus, for any triplet of region embeddings $g(w_{RA}), g(w_{RP}), g(w_{RN})$, the objective function to minimize is the triplet loss:
\begin{multline}
\mathcal{L}\big(g(w_{RA}), g(w_{RP}), g(w_{RN})\big) = \\
\max \Big(||g(w_{RA}) - g(w_{RP})||_2 \ - \\ 
||g(w_{RA}) - g(w_{RN})||_2 + m, 0\Big),
\end{multline}
where $m$ is a margin parameter. The margin parameter is meant to ensure sufficient separation between (anchor, positive) and (anchor, negative) pairs. In practice, the optimization does not stop until such pairs are at least a margin apart. In essence, the triplet loss function pushes the embeddings of geographically closer regions to be closer together in the embedding space while simultaneously pushing the embeddings of farther or dissimilar regions further apart.

In the experimental evaluation, we evaluate two different aggregators: one based on a transformer encoder, and the other based on a learnable weighted sum. More details on them are provided in Section \ref{sec: exp app methodology}.

\section{Experimental evaluation}
\label{sec:experiments}
To answer the research question introduced in Section \ref{sec: intro}, which we also formally expressed in Definition \ref{def: prob def}, Section \ref{sec: preliminaries}, we assess the effectiveness of our methodology for generating embeddings that capture salient features of urban regions. To this end, we conduct an experimental evaluation guided by the following questions:
\begin{itemize}
    \item \textbf{EQ1}: do the urban region embeddings generated by our methodology outperform those produced by a state-of-the-art method when evaluated on selected downstream tasks?
    \item  \textbf{EQ2}: do the embeddings capture the temporal characteristics and dynamics of urban regions?
%
%
\end{itemize}
%
In the following, we provide details on the datasets used in the evaluation (Section \ref{sec:datasets}), and how the proposed methodology was practically implemented (Section \ref{sec: exp app methodology}). Finally, Section \ref{sec:eq1} presents the experiments conducted to address \textbf{EQ1}, while Section \ref{sec:eq2} covers the experiments aimed at addressing \textbf{EQ2}.

\subsection{Datasets}
\label{sec:datasets}

In our experimental evaluation, we use four datasets: service-specific mobile traffic volume data provided by the NetMob 2023 challenge \cite{martínezdurive2023netmob23}, land use data from The Urban Atlas \cite{laduse2018dataset}, French census data provided by the IRIS French census cells, and POI data from Openstreetmap (via Geofabrik). Below, we describe each dataset in detail.

\vspace{0.3em}
\noindent \textbf{Service-specific mobile traffic volume data.} 
The NetMob 2023 dataset \cite{martínezdurive2023netmob23} contains high-resolution, multi-region, service-level mobile traffic data. It offers detailed information about the spatio-temporal traffic volume of mobile applications, covering 68 popular services. This data is geo-referenced with a resolution of $(100m)^2$ across 20 metropolitan areas in France, collected over 77 consecutive days in 2019, with a sampling rate of 15 minutes. In our experiments, we consider the city of Paris and aggregate the time series data hourly by summing the upload and download traffic.
From a total of 68 mobile services, we focus on those belonging to four selected macro categories that, we argue, reasonably represent a wide range of human behaviors and activities. These are: \textit{Social} (includes services such as Facebook, Instagram), \textit{Work} (e.g., Google Docs, email), \textit{Gaming} (e.g., Fortnite, Clash of Clans), and \textit{Streaming} (e.g., Netflix, Orange TV).

\vspace{0.3em}
\noindent \textbf{Landuse data.}
The \textit{Urban Atlas 2018} \cite{laduse2018dataset} is a high-resolution dataset that provides detailed and inter-comparable land use and land cover data for 788 Functional Urban Areas (FUAs) across several countries. The dataset also includes integrated population estimates for the 2018 reference year. It is based on satellite image interpretation, supplemented by data from Google Earth and OpenStreetMap. The Urban Atlas includes 23 land use and land cover categories, ranging from various types of urban fabric and industrial or commercial units to natural landscapes such as forests, pastures, and water bodies.

\vspace{0.3em}
\noindent \textbf{POI data.} We downloaded OpenStreetMap POI data, related to the Île-de-France region, via Geofabrik\footnote{\url{https://download.geofabrik.de/europe/france/ile-de-france.html}} We considered POIs, buildings and transport facilities, grouped by category.

\vspace{0.3em}
\noindent \textbf{IRIS data.}
The French National Institute of Statistics and Economic Studies (INSEE) established the IRIS (i.e., aggregated units for statistical information) system\footnote{\url{https://www.insee.fr/fr/metadonnees/definition/c1523}}, which partitions France into uniformly sized units. These units, referred to as IRIS units, are the primary means of disseminating infra-municipal data and are defined according to geographic and demographic criteria. Each unit must also have clearly identifiable and stable boundaries. IRIS units fall into three categories:
\begin{itemize}
\item \textit{Residential} units, with populations ranging from 1,800 to 5,000, are homogeneous in terms of living conditions. Their boundaries are marked by significant urban features, such as main roads, railways, or water bodies.
\item \textit{Business} units contain more than 1,000 employees, with a ratio of employees to residents of at least 2:1.
\item \textit{Miscellaneous} units cover large, sparsely populated areas such as leisure parks, ports, or forests.
\end{itemize}
Using the characteristics provided by the IRIS dataset, we compute population density by dividing the total number of residents (originally segmented by age groups) by the area of the IRIS unit in km$^2$.

\subsection{Application of our methodology to the NetMob 2023 dataset}
\label{sec: exp app methodology}

In the experimental evaluation, we aim to generate urban region embeddings for selected urban areas from the NetMob 2023 dataset. To achieve this, we instantiate the methodology proposed in Section \ref{sec:methodology}. 

Recall that we consider each $(100m)^2$ NetMob cell as a mobile traffic cell (MTC, Definition \ref{def: mtc}), while the time series associated with these cells correspond to the cellular time series (CTS) introduced in Definition \ref{def: stvts}.
%
%
We thus begin 
by preprocessing the original NetMob 2023 dataset, where we reduce the size of the dataset to decrease the computational and memory requirements of the autoencoder step (i.e., step 1). More precisely, for each considered CTS in the original dataset, we first downsample, reducing the sampling rate to 1 hour. This is followed by a category-wise aggregation, where the selected services are grouped into four macro categories: \textit{Social}, \textit{Work}, \textit{Gaming}, and \textit{Streaming}. This results in a set of multivariate aggregated time series, one per MTC (NetMob cell).

Next, in \textbf{step 1}, we train an autoencoder to generate MTC embeddings. The autoencoder takes a multivariate aggregated time series for each MTC as input and produces the MTC's embedding as output. Our autoencoder is based on temporal convolutional networks (TCNs) following the architecture in \cite{thill2020time}, with minor modifications to produce a 44-dimensional embedding for each MTC. The autoencoder is trained using mean squared error loss, with a learning rate of $10^{-3}$ over 100 epochs.

Finally, in \textbf{step 2}, we \textit{aggregate} the MTC embeddings into urban region embeddings: this requires to choose a target tessellation, as per Definition \ref{def: target partitioning} and the problem definition introduced in Definition \ref{def: prob def}.  The target tessellation we consider is based on the French IRIS system, as described in Section \ref{sec:datasets}, meaning that each region corresponds to an IRIS cell. If a region contains more than 300 MTCs, we randomly sample 300 of them to reduce the aggregator's computational cost. We report that in Paris, only 29 out of 2841 IRIS intersect more than 300 NetMob MTCs, hence making the loss of information negligible. 
For each region, we build a feature matrix $X$ of size $300 \times 44$, where each row contains an MTC embedding from step 1. If a region contains fewer than 300 MTCs, the remaining rows are padded with zeros. We also associate a mask matrix to indicate which values are padded.
We report that in step 2, we use and compare two different aggregators: 
\begin{itemize}
    \item \textbf{Learnable weighted sum}: defined as 
    $X'=\text{linear}_2(X\cdot\sigma(\text{linear}_1(X)))$,    
    where $\text{linear}_1,\ \text{linear}_2$ are two linear layers and $\sigma$ is the sigmoid. Produces embeddings of size $64$.
    
    \item \textbf{Transformer encoder with average pooling}: 
    we concatenate two encoder layers from the vanilla transformer \cite{vaswani2017attention}, each with a single attention head, followed by an average pooling layer. Produces embeddings of size 64.
\end{itemize}

Both aggregation methods are trained using the triplet loss defined in Section \ref{sec:methodology}, with the Adam optimizer, a learning rate of $10^{-4}$, and 60 training epochs. The contrastive learning relies on the concepts of anchor, positive, and negative examples. The degree of locality is controlled by the \textit{hops} parameter, which defines how far a region can be considered a positive example. For instance, hops = 1 means only regions directly adjacent to the anchor are positive, while hops = 2 includes regions adjacent to both the anchor and its neighbors, and so on.
The code implementing our methodology and experimental evaluation is written in Python and has been made available in a GitHub repository\footnote{Omitted for blind review.}.

\subsection{Addressing \textbf{EQ1}: evaluating the quality of embeddings}
\label{sec:eq1}

\begin{table*}[t]
    \caption{Land use inference in Paris.
    }\label{tab:landuse}
    \centering
    \begin{tabular}{lcccc}
    \toprule
    \textbf{Model} & \textbf{Category} & \textbf{KL-div} & \textbf{L1} & \textbf{Cosine Similarity} \\
    \midrule
    Our method (transformer) & Gaming & 0.4097$\pm$0.00719 & 0.57786$\pm$0.00951 & 0.86929$\pm$0.00324  \\
    Our method (weighted sum) & Gaming & 0.39468$\pm$0.00306 & 0.57096$\pm$0.00659 & 0.8779$\pm$0.0.0166  \\
    RegionDCL & -- & 0.35442$\pm$0.00497 & 0.53953$\pm$0.00804 & 0.88827$\pm$0.0027  \\
    Our method (weighted sum) & Work & 0.36034$\pm$0.00331 & 0.52982$\pm$0.00639 & 0.89068$\pm$0.00225  \\
    Our method (transformer) & Social  & 0.36256$\pm$0.0049 & 0.54108$\pm$0.00843 & 0.89351$\pm$0.00297  \\
    Our method (transformer) & Work  & 0.3481$\pm$0.00344 & 0.52313$\pm$0.00792 & 0.89389$\pm$0.00207  \\
    Our method (transformer) & Streaming  & 0.35147$\pm$0.00543 & 0.52538$\pm$0.00739 & 0.8967$\pm$0.00317  \\
    Our method (weighted sum) & Social & 0.36217$\pm$0.00402 & 0.53786$\pm$0.00746 & 0.8978$\pm$0.00205  \\
    Our method (weighted sum) & Streaming & 0.34865$\pm$0.00346 & 0.52264$\pm$0.00648 & 0.90046$\pm$0.0.0153  \\
    Our method (weighted sum) & All & 0.32842$\pm$0.01858 & 0.50268$\pm$0.01642 & 0.9036$\pm$0.00792  \\
    \textbf{Our method (transformer)} & \textbf{All} & \textbf{0.31268$\pm$0.01683} & \textbf{0.48522$\pm$0.01571} & \textbf{0.90641$\pm$0.00763} \\
    \bottomrule
    \end{tabular}
\end{table*}

\begin{table*}[t]
\caption{Population density estimation in Paris.}
\label{tab:pop}
\centering
\begin{tabular}{lcccc}
\toprule
\textbf{Model} & \textbf{Category} & \textbf{MAE} & \textbf{RMSE} & $\mathbf{R^2}$ \\
\midrule
RegionDCL & -- & 3627.7858$\pm$69.90744 & 5098.71439$\pm$77.67808 & 0.41924$\pm$0.01974 \\
Our method (transformer) & Gaming & 2952.01801$\pm$49.27191 & 4423.04382$\pm$98.42791 & 0.56283$\pm$0.02126 \\
Our method (weighted sum) & Gaming & 2763.29038$\pm$55.68725 & 4224.42716$\pm$95.46748 & 0.60116$\pm$0.0204 \\
Our method (transformer) & Social & 2759.55168$\pm$59.26227 & 4195.68068$\pm$93.38544 & 0.60665$\pm$0.01805 \\
Our method (weighted sum) & Work & 2702.95954$\pm$62.3961 & 4101.72277$\pm$95.62261 & 0.62407$\pm$0.01808 \\
Our method (transformer) & Streaming & 2538.08503$\pm$67.52267 & 3914.71953$\pm$95.78023 & 0.65752$\pm$0.01755 \\
Our method (transformer) & Work & 2597.41923$\pm$57.56329 & 3884.87063$\pm$75.48345 & 0.66283$\pm$0.01309 \\
Our method (weighted sum) & Social  & 2531.73449$\pm$80.44075 & 3839.25659$\pm$116.40961 & 0.67046$\pm$0.02108 \\
Our method (weighted sum) & Streaming  & 2383.41097$\pm$53.35184 & 3694.00752$\pm$93.69873 & 0.69503$\pm$0.01625 \\
Our method (transformer) & All & 2315.95861$\pm$86.13854 & 3653.39424$\pm$155.14478 & 0.70128$\pm$0.02599  \\
\textbf{Our method (weighted sum)} & \textbf{All} &\textbf{ 2243.77616$\pm$80.1004} & \textbf{3598.12695$\pm$189.25283} & \textbf{0.70987$\pm$0.03306}  \\
\bottomrule
\end{tabular}
\end{table*}

To address \textbf{EQ1}, we evaluate the region embeddings generated by our methodology using two supervised downstream tasks considered in many prior works in the literature: land use inference and population density estimation. We compare our methodology, instantiated as detailed in Section \ref{sec: exp app methodology}, with RegionDCL \cite{li2023urban}, a recent state-of-the-art approach that generates urban region embeddings from points of interest and building data. Additionally, we present the results for the two cell aggregators used in step 2 of our methodology, considering both individual service categories and all categories combined. Furthermore, we evaluate how well our embeddings perform in an unsupervised clustering task, again compared to the state-of-the-art.

\vspace{0.3em}
\noindent \textbf{Land use inference.}
The land use inference task is a label distribution learning problem. To this end, we use the region representations generated by both our methodology and RegionDCL to infer the proportion of land use categories within each region.
For the land use inference task, we employ a 2-layer Multi-Layer Perceptron (MLP) with 512 hidden units and four output units. The evaluation metrics used are L1 distance, KL-divergence, and Cosine similarity. We split the regions randomly into 70\% training, 10\% validation, and 20\% test sets. The MLP is optimized using KL-divergence loss for 100 epochs, with early stopping applied using a patience of 10 epochs. Each experiment is repeated 30 times, and the average values and standard deviations are reported in Table \ref{tab:landuse}.

As shown in Table \ref{tab:landuse}, our methodology, with both cell aggregators, outperforms RegionDCL. Although the transformer aggregator performs slightly better than the learnable weighted sum, the performance of both approaches is comparable. The best configuration for the transformer aggregator is with 3 hops, while for the learnable weighted sum, the best results occur with 2 hops. Increasing the number of hops (up to 5) decreases performance, which aligns with Tobler's first law of geography \cite{tobler1970computer}. Additionally, the results show that performance improves when considering all service categories together rather than individual categories. The \textit{Gaming} service produces the lowest performance, likely due to its lower traffic volume compared to categories like \textit{Social}, which may provide more distinctive information for this downstream task.

\vspace{0.2em}
\noindent \textbf{Population density estimation.} 
The population density estimation task is formulated as a regression problem. Similar to the land use inference task, we compare the performance of our embeddings with those generated by RegionDCL. The evaluation metrics used are mean absolute error (MAE), root mean squared error (RMSE), and the coefficient of determination ($R^2$). For this task, we randomly partition the datasets into 80\% training and 20\% test sets. We train a random forest regressor \cite{liaw2002classification} on 30 different data splits, reporting the average and standard deviation of the results in Table \ref{tab:pop}.

As shown in Table \ref{tab:pop}, our methodology, with both cell aggregators, again outperforms RegionDCL. While the learnable weighted sum-based aggregator slightly outperforms the transformer for this task, the performance of both remains comparable. The optimal configuration for both aggregators occurs with 2 hops. 
Similarly to the previous downstream task, we observe that aggregating all service categories leads to better performance. Notably, in this task, the \textit{Gaming} category performs slightly better, and the \textit{Work} category appears more informative for this task than for the land use inference task. The \textit{Social} and \textit{Streaming} services provide the best results, likely due to their higher traffic volumes and stronger relevance to the population density estimation task.

\vspace{0.2em}
\noindent \textbf{Clustering evaluation.}
Using embeddings to represent urban regions enables to measure similarities between them, thereby revealing common patterns. By applying a clustering algorithm, we can identify regions with comparable characteristics and, in turn, infer their functions, such as residential, business, or commercial areas, in a more data-driven manner. In this evaluation, we aim to assess how clustering urban regions based on embeddings generated from mobile traffic data compares to clustering them using traditional urban features such as POIs or land use. 
To this end, we employ hierarchical clustering using Ward’s method with squared Euclidean distance on the embeddings generated by the transformer-based MTC aggregator. We experiment with various cluster sizes $k$. Then, we also apply the same clustering method to the above-mentioned land use and POI data.

%

To assess the similarity between clusterings, we use the Adjusted Mutual Information (AMI) index, which measures the similarity between two clusterings based on different features \cite{vinh2009information}. In this context, the AMI score reflects the level of agreement between clusterings derived from service-specific mobile traffic data and traditional urban features. This approach can be considered a downstream task applied in an unsupervised setting.
Before presenting the results, we briefly go over the definition of AMI score.
Let $\mathbf{U}=\{u_1,u_2, \ldots,u_m\}$ be a clustering of a set $A$. 
The \textit{entropy} of $\mathbf{U}$ is defined as:
$$H(\mathbf{U}) = -\sum_{i=1}^m p(u_i) \log (p(u_i)),$$
where $p(u_i)$ is the probability that a point in $A$ belongs to cluster $u_i$. Intuitively, the entropy represents the uncertainty in clustering $\mathbf{U}$ and is non-negative, reaching 0 only when $\mathbf{U}$ contains a single cluster.
Let then $\mathbf{V} =\{v_1,v_2, \ldots,v_n\}$ be a different clustering of the same set $A$.  The \textit{mutual information} between $\mathbf{U}$ and $\mathbf{V}$ is defined as:
$$I(\mathbf{U},\mathbf{V}) = \sum_{i=1}^m \sum_{j=1}^n  p(u_i,v_j) \log \frac{p(u_i,v_j)}{p(u_i)p(v_j)},$$
where $p(u_i,v_j)$ is the probability of a point belonging to both $u_i$ and $v_j$. Mutual information quantifies how much information is shared between the two clusterings. It is zero when the two clusterings are independent and maximized when $\mathbf{U} = \mathbf{V}$.

The \textit{adjusted mutual information index} is a variation of mutual information that addresses the problem of clustering dependencies that arise by chance. It is defined as: 
$$\text{AMI}(\mathbf{U},\mathbf{V})= \frac{I(\mathbf{U},\mathbf{V})-\mathbb{E}(I(\mathbf{U},\mathbf{V}))}{\max \{H(\mathbf{U}),H(\mathbf{V})\}-\mathbb{E}(I(\mathbf{U},\mathbf{V}))},$$ 
where $\mathbb{E}(I(\mathbf{U},\mathbf{V}))$ is the expected mutual information between two random clusterings.
In traditional clustering, probabilities $p(u)$ are computed as the number of elements in $u$ divided by the size of set $A$. However, in the case we are considering in this work, to account for the size of each urban region we calculate $p(u)$ as the ratio of the total area of $u$ over the total area of $A$.

Table~\ref{tab:landuse_poi_results} presents the AMI scores comparing the clusterings obtained from the embeddings generated by our method (using the transformer as the aggregator) with those based on land use and POI data. The results shown in the table account for different numbers of clusters.
For comparison, we also evaluate the clustering derived from RegionDCL embeddings using the same method. Note that higher AMI scores indicate greater alignment, reaching a maximum value of 1 for identical clusterings and an expected value of 0 for random clusterings. 
\begin{table}[t]
\centering
\caption{AMI scores comparing clusterings between our/RegionDCL urban representations and traditional urban features from land use and POIs.}
\label{tab:landuse_poi_results}
\begin{tabular}{lccc}
\toprule
\textbf{k} & \textbf{Method} & \textbf{Land use} & \textbf{POIs} \\ 
\midrule
\multirow{2}{*}{5} & RegionDCL & 0.0637 & 0.0495 \\
                    & Our method (transformer) & \textbf{0.1199 }& \textbf{0.0720} \\ 
                    
\midrule
\multirow{2}{*}{10} & RegionDCL & 0.1312 & 0.0848 \\ 
                    &  Our method (transformer) & \textbf{0.1485} & \textbf{0.1130 }\\ 
                    
\midrule
\multirow{2}{*}{15} & RegionDCL & 0.1496 & 0.0997 \\ 
                    &  Our method (transformer) & \textbf{0.1867} & \textbf{0.1309} \\ 
                    
\midrule
\multirow{2}{*}{19} & RegionDCL & 0.1756 &0.1215\\ 
                    &  Our method (transformer) & \textbf{0.2049} & \textbf{0.1492} \\ 
                    
\bottomrule
\end{tabular}
\end{table}

The results demonstrate a stronger agreement between clustering the urban regions based on mobile traffic embeddings and traditional urban features compared to the state-of-the-art RegionDCL embeddings. It is also noteworthy that the RegionDCL embeddings already incorporate POI data, making the higher AMI scores achieved by our method even more significant.

\vspace{0.3em}
\noindent \textbf{Discussion.} The results presented in this section address \textbf{EQ1} and demonstrate that our methodology consistently outperforms RegionDCL across all tasks. This indicates that service-specific mobile traffic data can be effectively leveraged to generate high-quality urban region embeddings that rival and often surpass, those produced from more traditional, static data sources. Furthermore, the findings show that combining multiple service categories leads to consistently better-quality embeddings. Finally, the clustering evaluation reveals that embeddings generated by our methodology capture more urban information when grouping regions in an unsupervised manner compared to those of RegionDCL.

\subsection{Addressing \textbf{EQ2}: evaluation of temporal dynamics}
\label{sec:eq2}

To address \textbf{EQ2}, we aim to demonstrate that the use of service-specific mobile traffic data allows us to capture the temporal evolution of city regions throughout the day.
%
%
Indeed, we argue that this is made possible by the highly dynamic nature of the data we are considering, which can better capture the temporal dynamics of urban areas than static data sources such as POIs. 

Specifically, we partition the time series in the NetMob 2023 dataset into three time slots based on the timestamp: \textit{night} (00:00 to 07:59), \textit{morning} (08:00 to 15:59), and \textit{afternoon} (16:00 to 23:59), and compute embeddings for each time slot. By incorporating the four macro categories of mobile services described in Section \ref{sec:datasets}, we aim to effectively describe how regions change during the day. 

We cluster urban regions considering the embeddings of the three time slots and full-day time series adopting the same clustering method described in Section \ref{sec:eq1}. 
To select the best number of clusters, we apply the Elbow method to the inertia curve, together with the study of the silhouette curve: we found out that $k = 9$ is the best across all the embeddings.
\begin{figure}[t]
    \centering
    \includegraphics[width=0.9\linewidth]{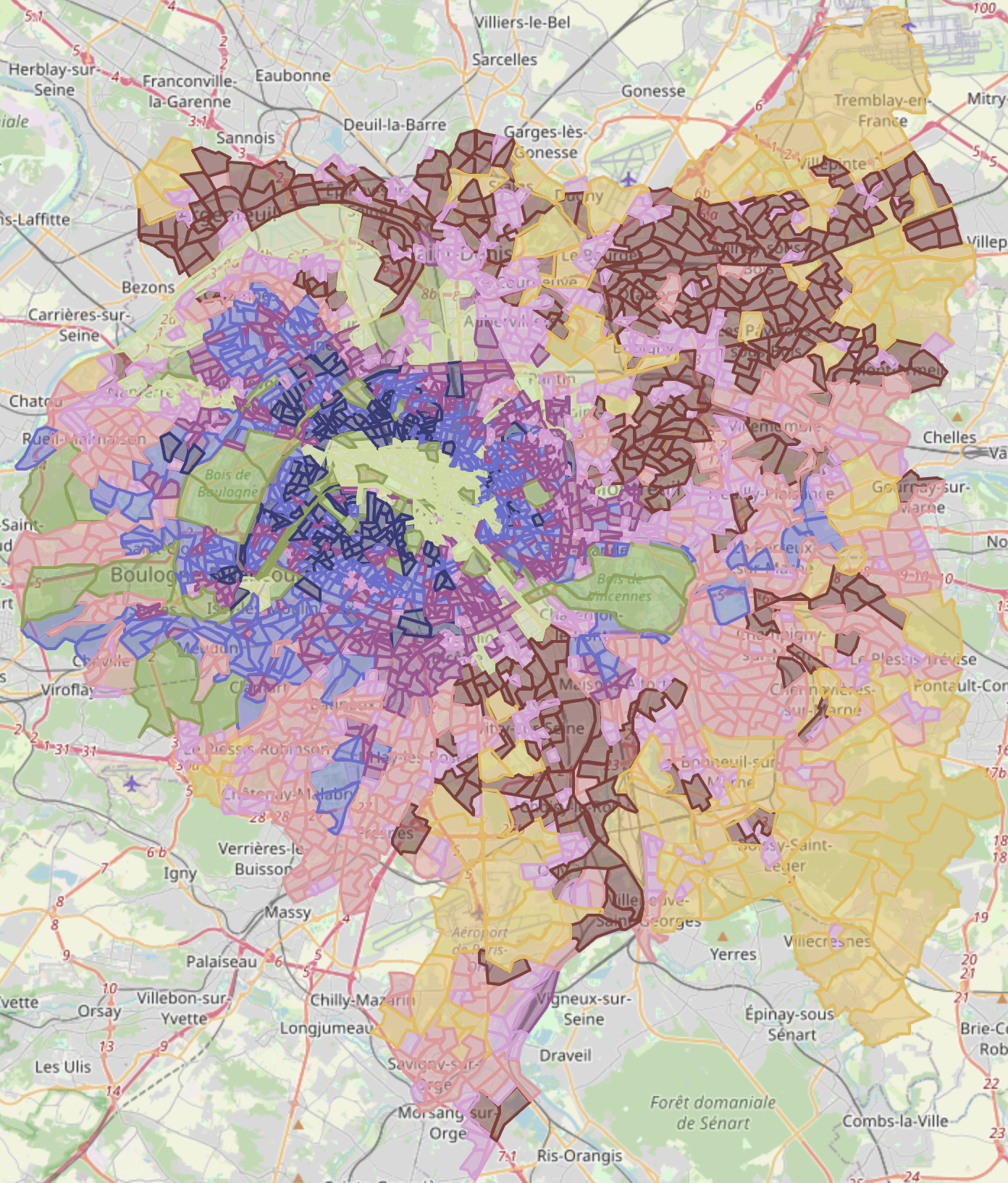}
    \caption{Clustering on full-day time series embeddings ($k=9$). }
    \label{fig:full-day}
\end{figure}

\begin{figure*}[ht]
    \centering
     \subfloat[Morning]{\includegraphics[width=.32\textwidth]{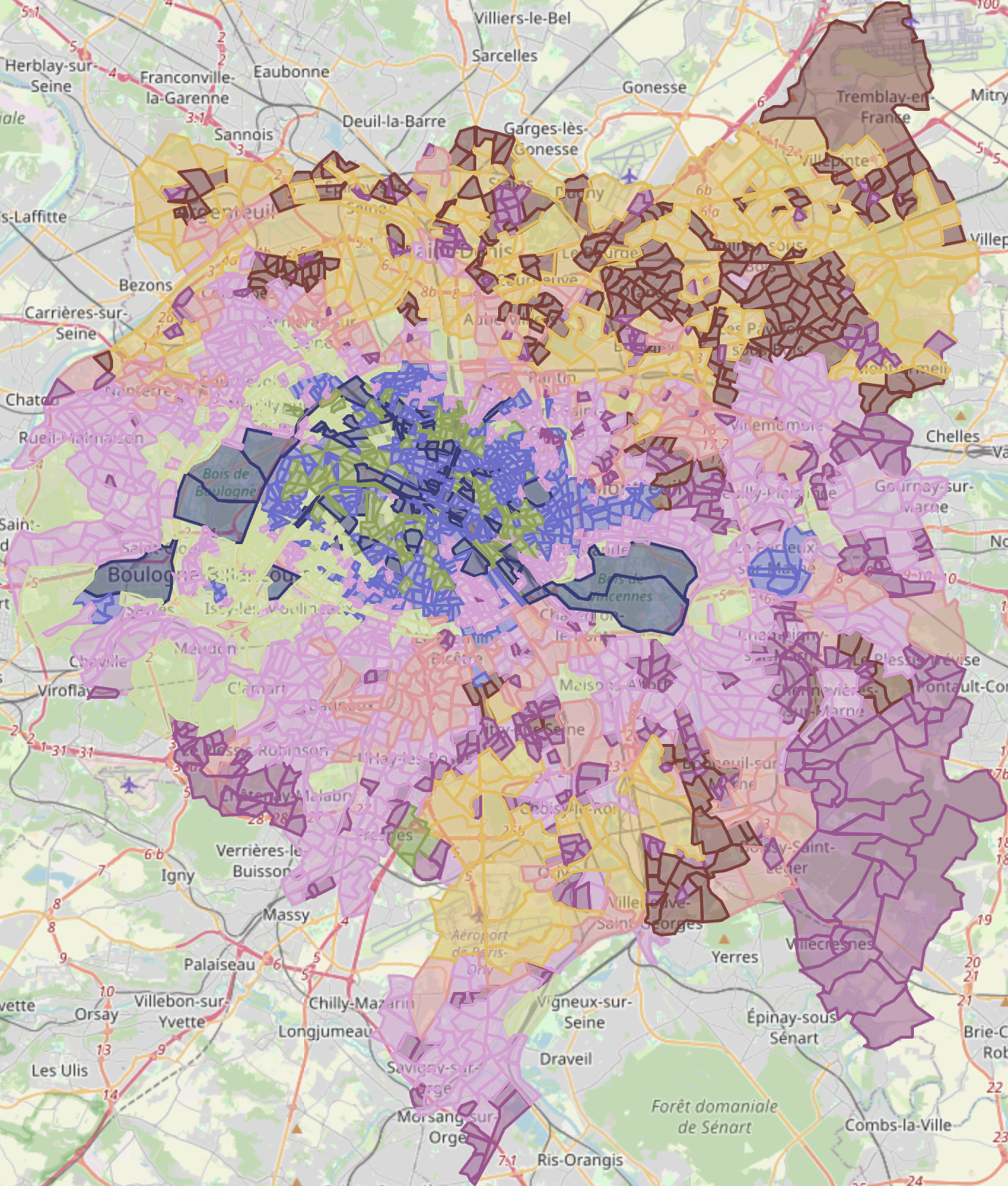}\label{fig:morning}} \hspace{.1mm}
     \subfloat[Afternoon]{\includegraphics[width=.32\textwidth]{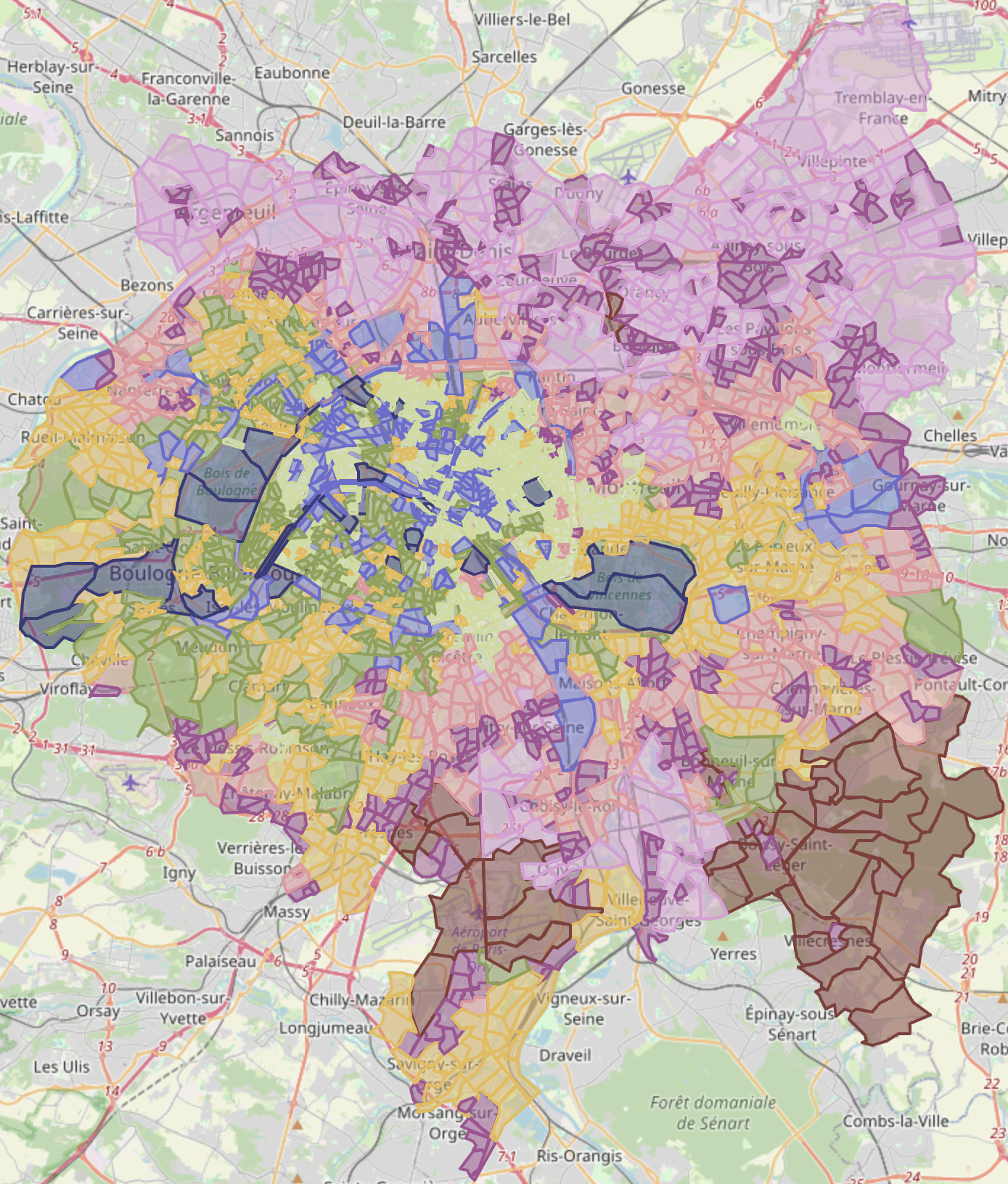}\label{fig:afternoon}} \hspace{.1mm}
     \subfloat[Night]{\includegraphics[width=.32\textwidth]{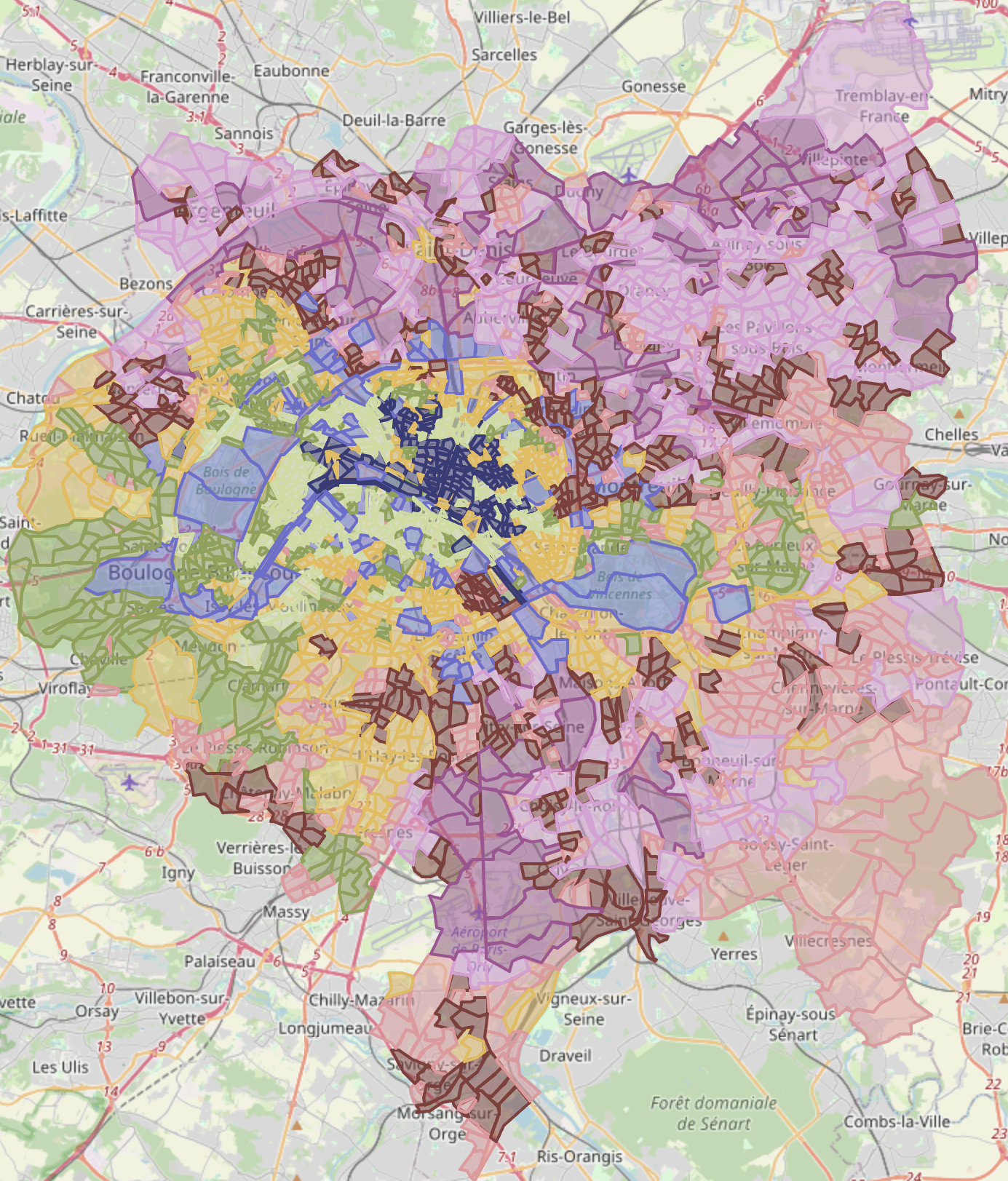}\label{fig:night}}
     \caption{Clustering of the embeddings of the three time slots series ($k=9$). }\label{fig:clustering}
\end{figure*}

Analyzing the resulting clusters, we first observe that, for the full-day time series (see Figure \ref{fig:full-day}), parks and major landmarks such as the cemeteries, Champs-Élysées, and the Eiffel Tower consistently cluster together, indicating shared traffic dynamics likely driven by recreational and tourist activities. This grouping remains stable during the morning (dark blue cluster in Figure \ref{fig:morning}), but during the afternoon and night (dark blue and violet cluster in Figures \ref{fig:afternoon} and \ref{fig:night} respectively) some central locations, like the Louvre, shift to another cluster, possibly reflecting different traffic patterns during these times. Particularly, during the night, we observe that this cluster also includes regions close to the Seine river. It is worth noticing that, even if our methodology generates embeddings from service-specific mobile traffic data, it is able to cluster green areas together, similarly to what happens with embeddings produced by methods in other works that use POI data as input.

Regarding the airports, a pattern emerges where Paris Orly often clusters together with the smaller Paris Le Bourget airport, particularly during the morning and night, while Paris Charles de Gaulle airport remains in a separate cluster (see Figures \ref{fig:morning} and \ref{fig:night}). This suggests distinct activity levels, with Paris Charles de Gaulle maintaining different traffic, especially during the night when Paris Orly and Paris Le Bourget group together. In the full-day analysis, however, all airports appear in the same cluster, reflecting overall connectivity and similar usage patterns when averaged throughout the day (see Figure \ref{fig:full-day}). 

For what concerns green areas, such as forests and parks, this is particularly interesting because, even without using POIs or similar data as input to our methodology, we are still able to group regions with similar characteristics. Moreover, by observing the temporal evolution of the region embeddings, we can see how these areas change throughout the day. This might be used, for instance, to optimize public transportation by increasing service to one airport rather than another during specific time slots.

The Musée d’Orsay exhibits a unique behavior by consistently clustering with more suburban residential areas, rather than the neighboring zones, during the morning and afternoon. This  indicates a different traffic profile from other central museums or attractions (see Figures \ref{fig:morning} and \ref{fig:afternoon}). However, in the full-day analysis shown in Figure \ref{fig:full-day}, the museum shifts towards more central clusters, though still separate from major attractions like the Louvre. Also, in this case, we can take advantage of the time ranges and use this information to better understand how people behave in that area and investigate why the behavior is different in that region during the morning and afternoon. 

Another example concerns the train stations, which show varying behavior depending on the time range. Some are grouped together, particularly at night when all stations form a central cluster, reflecting a more unified traffic pattern during lower-traffic hours (Figure \ref{fig:night}). In contrast, during the morning and full-day analysis, the stations are more scattered across clusters, suggesting different traffic patterns in various parts of the city (Figures \ref{fig:full-day} and \ref{fig:morning}). 

Finally, we highlight some differences when we consider the full-day clustering and the clusterings obtained with the three time slots. The full-day clustering presents a large cluster in the northern part of the city center, suggesting a major convergence of activity in that area when considering the entire day and, consequently, less granularity in recognizing differences between regions in the city center (see light green cluster in Figure \ref{fig:full-day}). 

In conclusion, the qualitative analysis we conducted demonstrates that the use of service-specific traffic data as input data to generate urban region embeddings allows to answer question \textbf{EQ2}, as the embeddings seem to effectively capture region temporal dynamics.
This is just one example of how this data can be used to understand a city's temporal dynamics. Going back to the previous example of improving and intensifying transportation in specific regions and time slots, one could increase the granularity of these intervals for greater precision. At the same time, exploring the differences between weekdays, weekends, and holidays could provide deeper insights into traffic patterns.







\section{Conclusions and Future Works}
\label{sec: conclusions}

The recent availability of service-specific mobile traffic data has opened up new opportunities for exploiting such data beyond the management of services and networking towards new applications, like urban analysis. In this paper, we introduce a novel methodology that leverages time series data from mobile service traffic to capture both static and dynamic characteristics of urban regions. Our primary research question investigates the effectiveness of service-specific mobile traffic data in representing key features of urban areas. The proposed methodology involves two main steps: first, generating cellular embeddings from service-specific multivariate time series and then aggregating cells into regions using a contrastive learning approach. We tested our approach on a real-world dataset, evaluating two aspects: the quality of the resulting embeddings and their ability to capture regional dynamics. Extensive experiments demonstrate that service-specific mobile data effectively captures both the characteristics and dynamics of urban regions, highlighting its potential for a range of novel applications, emphasizing the importance of making such data accessible
to support public innovation.
Our future works follow different directions.
First, we aim to enhance the evaluation process of embedding quality by designing specific use cases. We will explore various strategies for generating embeddings, focusing on Graph Neural Network approaches that account for the urban graph structure. Additionally, we plan to investigate transfer learning techniques to estimate traffic volumes in previously unobserved areas, leveraging the established relationship between geographical context and mobile phone usage.


\noindent \textbf{Acknowledgments.}
The NetMob 2023 dataset has been made available by Orange within the framework of the NetMob 2023 Challenge \cite{martínezdurive2023netmob23}.

\clearpage

\bibliographystyle{IEEEtran}
\bibliography{biblio}

\end{document}